\documentclass[twoside,11pt]{article}

\usepackage{jmlr2e}
\usepackage[utf8]{inputenc} % allow utf-8 input
\usepackage[english]{babel}
\usepackage[T1]{fontenc}    % use 8-bit T1 fonts
\usepackage{hyperref}
\hypersetup{colorlinks,allcolors=black}
\usepackage{url}            % simple URL typesetting
\usepackage{booktabs}       % professional-quality tables
\usepackage{amsfonts}
\usepackage{nicefrac}       % compact symbols for 1/2, etc.
\usepackage{microtype}      % microtypography
\usepackage{xcolor}         % colors
\usepackage{mathrsfs}

\def\nn{\nonumber}
\def\beq{\begin{equation}}
\def\eeq{\end{equation}}
\def\bea{\begin{eqnarray}}
\def\eea{\end{eqnarray}}
\def\ba{\begin{array}}
\def\ea{\end{array}}

\def\bitem{\begin{itemize}}
\def\eitem{\end{itemize}}
\def\ben{\begin{enumerate}}
\def\een{\end{enumerate}}

\def\ie{{\it i.e.,\ \/}}

\definecolor{bgrd}{rgb}{1,1,1}
\definecolor{gray}{rgb}{0.5,0.5,0.5}
\definecolor{dkr}{rgb}{0.7,0.1,0.2}
\definecolor{dkb}{rgb}{0.1,0.1,0.8}

\makeatletter
\newdimen{\captionwidth}
\long\def\@makecaption#1#2{%
\captionwidth .9\hsize% use current value of \hsize
\vskip 10pt%
\setbox\@tempboxa\hbox{#1: #2}%
  \ifdim \wd\@tempboxa >\captionwidth%
    \setbox\@tempboxa\hbox{#1:\hspace*{.5em}}%
    \hfil\parbox{\captionwidth}{\raggedright\hangindent \wd\@tempboxa%
    \hangafter=1\unhbox\@tempboxa#2}\hfill%
  \else\centerline{\box\@tempboxa}%
  \fi
}
\makeatother
\def\scalefig#1{\epsfxsize #1\textwidth}

\def\T{\mbox{\tiny T}}

\newcommand{\mbbE}{\mathbb{E}}

\newcommand{\Fmsc}{\mathscr{F}}

\newcommand{\Pmsc}{\mathscr{P}}

\newcommand{\Xmsc}{\mathscr{X}}

\def\etabf{\hbox{\boldmath$\eta$\unboldmath}}

\def\nubf{\hbox{\boldmath$\nu$\unboldmath}}

\def\thetabf{{\mbox{\boldmath$\theta$\unboldmath}}}
\def\ubf{{\bf u}}

\def\xbf{{\bf x}}

\def\xbf{{\bf x}}

\def\Fbf{{\bf F}}
\def\Gbf{{\bf G}}
\def\Hbf{{\bf H}}

\def\Ac{{\cal A}}

\def\Dc{{\cal D}}

\def\Hc{{\cal H}}

\def\Uc{{\cal U}}

\usepackage{amsmath,amssymb,mathrsfs}
\usepackage{mathtools}

\usepackage{graphicx,stfloats,multicol}
\usepackage{epsfig,epsf,psfrag,amssymb,amsfonts,latexsym,graphicx,mathrsfs,subcaption}
\usepackage{bbm}
\usepackage{chngpage}
\usepackage{url}
\usepackage[hyphenbreaks]{breakurl}
\usepackage[normalem]{ulem}
\usepackage{algorithm}
\usepackage{algorithmic}

\newtheorem{Definition}{Definition}

\ShortHeadings{Innovations Autoencoder and its Application in One-class Anomalous Sequence  Detection}{Wang and Tong}
\firstpageno{1}

\begin{document}

\title{Innovations Autoencoder and its Application in \\One-class Anomalous Sequence  Detection}

\author{\name Xinyi Wang \email xw555@cornell.edu \\
       \addr Department of Electrical and Computer Engineering\\
       Cornell University\\
       Ithaca, NY 14850, USA
       \AND
       \name Lang Tong \email lt35@cornell.edu \\
       \addr Department of Electrical and Computer Engineering\\
       Cornell University\\
       Ithaca, NY 14850, USA}

\editor{}

\maketitle

\begin{abstract}%   <- trailing '%' for backward compatibility of .sty file
An innovations sequence of a time series is a sequence of independent and identically distributed random variables with which the original time series has a causal representation.  The innovation at a time is statistically independent of the  history of the time series.  As such, it represents the new information contained at present but not in the past.  Because of its simple probability structure, an innovations sequence is the most efficient signature of the original.  Unlike the principle or independent component analysis representations,  an innovations sequence preserves not only the complete statistical properties but also the temporal order of the original time series.  An long-standing open problem is to find a computationally tractable way to extract an innovations sequence of non-Gaussian processes.  This paper presents a deep learning approach, referred to as Innovations Autoencoder (IAE), that extracts innovations sequences using a causal convolutional neural network.   An application of IAE to the one-class anomalous sequence detection problem with unknown anomaly and anomaly-free models is also presented.
\end{abstract}


\section{Introduction}
\label{sec:intro}
At the heart of modern machine learning is the notion of   ``signature.''   A data signature should capture essential characteristics of the data and has low dimensionality for easy processing.  This work focuses on extracting signatures from a random process (time series) for real-time machine learning applications such as anomaly detection, target tracking, control, and system monitoring.

Assume that, at time t, we have observations $\Xmsc_t=\{x_t, x_{t-1},\cdots\}$ of a stationary random process $(x_t)$, and there are infinitely many data samples to arrive, one at a time, in the future.  We are interested in real-time statistical learning problems that make inference or control decisions based on $\Xmsc_t$ under the general framework that the underlying probability model of $(x_t)$ is partially or entirely unknown.    The real-time (or online) nature of such machine learning problems makes it essential to generate, {\em causally},  a signature sequence $(\nu_t)$ from past samples such that it captures all statistical information of $(x_t)$ up to time $t$.  Mathematically, we are interested in low dimensional but {\em sufficient} statistics with a simple probability structure.

 A classic example of an efficient signature is the {\em innovations process} introduced by \cite{Wiener:58Book}.  Wiener and Kallianpur  pioneered the idea of encoding and decoding of  a stationary ergodic process $(x_t)$  via  an independent and identically distributed ({\it i.i.d.}) sequence $(\nu_t)$ referred to as an {\em innovations sequence} by \cite{Masani66BAMS}.  In particular,  the {\em innovations representation} of $(x_t)$ is defined by an encoder $G$ that produces an {\it i.i.d.} sequence $(\nu_t)$ by a {\em casual transform}:
\beq \label{eq:G}
\nu_t = G(x_t,x_{t-1},\cdots),
\eeq
and a decoder $H$ that recovers $(x_t)$ from $(\nu_t)$ also via a causal transform:
\beq \label{eq:H}
x_t = H(\nu_t, \nu_{t-1},\cdots).
\eeq
The existence of such a representation notwithstanding,  the interpretation of $(\nu_t)$ as an {\em innovations} sequence of $(x_t)$ is immediate:  $\nu_t$,  being independent of $\Xmsc_{t-1}=(x_{t-1}, x_{t-2},\cdots)$, represents the new information  in $x_t$ but not in $\Xmsc_{t-1}$.    Equally important is the decoding process, which makes the innovations sequence a sufficient statistics for all decisions made based on $\Xmsc_t$.    From a modern machine perspective, the encoder-decoder construct of the innovations representation is simply a {\em causal autoencoder} with innovations sequence as the latent process.

The classical theory on innovations has a long and illustrious history, starting from the work of \cite{Kolmogorov:41}, \cite{Wiener:58Book}, and \cite{Kalman:60} on prediction, filtering, and control, and it soon became popular in the engineering community; see \cite{Kailath:70Proc,Kailath:74IT}.   The innovations approach is particularly powerful for statistical inference and decision making when the innovations process in (\ref{eq:G}) can be easily computed in real-time.  One prominent case is the stationary Gaussian process for which the innovations process $(\nu_t)$ is simply the error sequence of the linear minimum mean-squared error  (MMSE) predictor.  Another case is the continuous-time (possibly non-Gaussian) process $(x_t)$ under the additive white Gaussian noise model, for which the innovation process is the error sequence of the {\em nonlinear} MMSE predictor  \citep{Kailath:71BJST}.  In general, however, there is no computationally tractable way to extract an innovations sequence when an innovations representation exists. This paper aims to bridge this gap using modern machine learning techniques and demonstrate the potential of innovations representation in an open  challenge of one-class anomalous sequence detection.

\subsection{Summary of results}
We develop a deep learning approach, referred to as Innovations Auto-Encoder (IAE), that provides a practical way to extract innovations of discrete-time stationary random processes with unknown probability structures, assuming that historical training samples are available.    To this end, we propose a causal convolutional neural network, {\it a.k.a}  time-delayed neural network \citep{Waibel&etal:89}, and a Wasserstein generative adversary network (GAN) learning algorithm for extracting innovations sequences \citep{Goodfellow&Shlen&Szegedy:15,Arjovsky17}.  Because the implementation and training of an IAE involve only a finite dimensional data vectors,  a convergence property is needed that ensures a sufficiently high-dimensional implementation will lead to a close approximation of the actual innovations representation. Under ideal training and implementation conditions,  we establish a finite-block convergence property in Theorem~\ref{thm:converge}, which ensures that a sufficiently high-dimensional implementation of an IAE,  trained with finite dimensional historical data vectors,  produces a close approximation of the ideal innovations representation.

Next,  we apply the idea of IAE to a ``one-class'' anomalous sequence detection problem where neither the anomaly-free nor the anomaly model is known, but anomaly-free training samples  are given.   By {\em anomaly sequence detection},  we mean to distinguish the underlying probability distributions of the anomaly-free  model  and that of the anomaly.    Although there are many practical machine learning techniques  for outlier and out-of-distribution (OoD) detection, to our best knowledge,   the result presented here is the first one-class anomalous sequence detection approach for time series models  with unknown underlying probability and dynamic models.

The problem of detecting anomalous sequence brings considerable computational and learning-theoretic challenges.  Although one expects that taking a large block of consecutive samples can reasonably approximate statistical properties of a random process, applying existing  detection schemes
 to such high-dimensional vectors with unknown (sequential) dependencies among its components is nontrivial.   The main contribution of this work is leveraging of the innovations representation to transform the anomaly-free time series to a sequence of (approximately) uniform {\it i.i.d.} innovations, thus making the anomalous sequence detection problem the classic problem of testing uniformity, for which we apply versions of coincidence test \citep{David:50Biometrika,Viktorova:64TPA,Paninski:08TIT,Goldreich:17book}.   We then demonstrate, using field-collected and synthetic datasets, the effectiveness of the proposed approach on detecting system anomalies in a microgrid \citep{Pignati&etal:15PESISG}.

\subsection{Notations}
Notations used are standard. All variables and functions are real.  We use  $\mathbb{R}^m$ and $\mathbb{N}$  for the $m$-dimension of real vector space and the set of integers, respectively.  Vectors are in boldface, and $\xbf=(x_1,\cdots, x_m) \in \mathbb{R}^m$  is a {\em column vector}.  A time series is denoted as $(x_t)$  with $t \in \mathbb{N}$.  Denote by  $\xbf_t^{(m)}:=(x_t, \cdots, x_{t-m+1})$ the column vector of current and $(m-1)$ past samples of $(x_t)$.

Suppose that  $F$ is an $m$-variate scaler function.  Let $\Fbf^{(n)}: \mathbb{R}^{m+n-1} \rightarrow \mathbb{R}^n$ be the $n$-fold time-shifted mapping of $F$ defined by $\Fbf^{(n)}(\xbf_t^{(n+m-1)}) := (F(\xbf_t^{(m)}), \cdots F(\xbf_{t-n+1}^{(m)}))$.   We drop the superscripts when the dimensionality is immaterial or obvious from the context.

\section{Background and Related Work}
\label{sec:background}

\subsection{Innovations representation and estimation}  We refer the readers to \cite{Kailath:70Proc} for an exposition of historical developments of the innovations approach. In general, an innovations representation of a stationary and ergodic process may not exist.  The existence of a causal encoder that maps $(x_t)$ to a uniform $i.i.d$ sequence holds quite generally for a large number of popular nonlinear time series models \citep{Wiener:58Book, Rosenblatt:59,Rosenblatt:09,Wu:05PNAS,Wu11:SI}.  The existence of a causal decoder that recovers the original sequence from an innovations sequence requires additional assumptions \citep{Rosenblatt:59,Rosenblatt:09,Wu11:SI}.  Whereas general conditions for the existence of an innovations representation are elusive, a relaxation on the requirement for the decoder  to produce a random sequence $(\hat{x}_t)$ with the same conditional (on past observations) distributions as that of $(x_t)$ makes the innovations representation applicable to a significantly larger class of practical applications as shown in \cite{Wu:05PNAS,Wu11:SI}. In this paper, we shall side-step the question of the existence of an innovations representation and focus on  learning the innovations representation $(H, G)$ in (\ref{eq:G}-\ref{eq:H}) when such a representation does exist.

Although there are no known ways to extract (or estimate) innovations when the underlying probability model is unknown,  several existing  techniques can be tailored for this purpose.  One way is to estimate $(\nu_t)$ by the error sequence of a linear or nonlinear MMSE predictor, which can be implemented and trained using a causal convolutional neural network (CNN). Such an approach can be motivated by viewing $(x_t)$ as a sampled process from a {\em continuous-time process} $\tilde{x}(t) = \tilde{z}(t) + w(t)$ in some interval,   where $w(t)$ is a white Gaussian noise and $z(t)$ a  possibly non-Gaussian but strictly stationary process.  Under mild conditions \citep{Kailath:71BJST}, the continuous-time innovations process $\tilde{\nu}(t)$ of $\tilde{x}(t)$ turns out to be the MMSE  prediction error process.  Unfortunately, there is no guarantee that the discrete-time version of the nonlinear MMSE  predictor will be an innovation sequence for $(x_t)$.

If we ignore the requirement that the innovation process $(\nu_t)$  needs to be a {\em causally invertible transform} of $(x_t)$, one can view the innovations sequence  $(\nu_t)$ as independent components of $(x_t)$,  for which there is an extensive literature since the seminal work  of \cite{Jutten&Herault:91} and \cite{Comon:94} on independent component analysis (ICA).  Originally proposed for linear models, ICA is a generalization of the principal component analysis (PCA) by enforcing statistical independence on the latent variables.  A  line of approaches akin to modern machine learning is to pass $x_t$ through a nonlinear transform to obtain an estimate of an {\it i.i.d.} sequence $(\nu_t)$, where the nonlinear transform can be updated based on some objective function that enforces independence conditions.  Examples of objective functions include information-theoretic and higher-order moment based measures \citep{Comon:94,Karhunen&etal:97NN,Naik&Kumar}.

 The ICA approach most related to this paper is ANICA proposed by  \cite{Brakel&Bengio:17}, where a  deep learning approach to nonlinear ICA via an autoencoder is trained by the Wasserstein GAN  \citep{Arjovsky17} technique. The main difference between IAE  and ANICA  lies in how causality and statistical independence are enforced in the learning process.  Different from IAE, ANICA does not enforce causality in training and its implementation.  It achieves statistical independence among extracted components through repeated re-samplings.

Also relevant is  NICE  \citep{DinhKruegerBengio2015} where a class of bijective mappings with unity Jacobian is proposed to transform blocks of $(x_t)$ to Gaussian  {\it i.i.d.} components.  The property of unity Jacobian makes NICE a particularly attractive architecture capable of evaluating relevant likelihood functions for real-time decisions.  However,  the special form of bijective mappings destroys the causality of the data sequence, and the requirement of {\it i.i.d.} training samples in the maximum-likelihood-based learning is difficult to satisfy for time series models.

It is natural to cast the problem of extracting independent components as one of designing an autoencoder where the latent variables are constrained to be statistically independent.  Several variational auto-encoder (VAE) techniques  \citep{KingmaWelling14,Kingma17,Tucker2018,MaaloeEtal19,VahdatKautz21}   have been proposed to produce generative models for the observation process using independent latent variables.  Without requiring that the latent variables are directly encoded from and capable of reproducing the original process, these techniques do not guarantee that the latent independent components are part of an innovations sequence. 

Unlike the non-parametric approach to obtaining innovations representations considered in this work, there is the literature on parametric techniques to extract innovations by assuming that  $(x_t)$ is a causal transform of an innovations sequence $(\nu_t)$.  By identifying the transform parameters, one can construct a causal inverse of the transform (if it exists).  From this perspective, extracting innovations can be solved in two steps: estimating first  the parameter estimation of a time series model, and (ii) constructing a causal inverse of the time series model.  Under relatively general conditions, parameters of multivariate moving-average and auto-regressive moving average models can be learned by moment methods involving high-order statistics \citep{Cardoso:89ICASSP,Swami&Mendel:92TAC, Tong:96SP}.

\subsection{One-class anomalous sequence detection}
The one-class anomalous sequence detection problem\footnote{In this paper, we do not consider the broader class of one-class anomaly detection problems where unlabeled training data or scarcely labeled anomalous training data are also used.} is a special instance of semi-supervised  anomaly detection that classifies a data sequence as anomalous or anomaly-free
 when training samples are given only for the anomaly-free model.   
To our best knowledge, there is no machine learning techniques specifically designed for time series, although there is an extensive literature on the related problem of  detecting outlier or OoD samples. 

A well know technique for  the one-class anomaly detection is the the one-class support vector machine (OCSVM)  \citep{Scholkopf:99NIPS} and its many variants for different applications \citep{Khan&Madden:04}. OCSVM finds the decision region for the anomaly-free samples by fencing in training samples with a ceratin margin that takes into account unobserved anomaly-free samples.  A related idea is  to separate the anomaly and anomaly-free model in the latent variable space of an autoencoder.  One such technique is  f-AnoGAN proposed by \cite{Schlegl&Seebock:19} where the OoD detection is made by fencing out all samples that result in large decoding error by an autoencoder trained with anomaly-free samples.  Other similar techniques include \citep{Bergmann19,GongEtal2019}.  
These discriminative methods rely,  {\em implicitly},  on an assumption that the anomaly  distribution complements that of the anomaly-free, \ie  the anomaly distribution concentrates in the region that the anomaly-free training samples do not or less likely to appear.
Therefore, they often under or over generalize the anomaly-free model and  perform  poorly when the domains of anomaly and anomaly-free models overlap completely.

Another set of techniques attach some kind of confidence scores on samples in the feature space, leveraging the neural network's ability  to learn the posterior distribution of the anomaly-free model  \citep{Hendrycks&Gimpel:17ICLR,Lakshminarayanan&Pritzel&Blundell:17NIPS,Liang&Li&Srikant:18ICLR,Lee&etal:18NIPS}.
These techniques construct confidence scores from the learned  anomaly-free model without attempting to learn or infer the anomaly model. As shown in \citep{Lan&Dinh:21}. even with the perfect density estimate, OoD detection may still perform poorly.

The third type of techniques simulate OoD samples in someway, often around the anomaly-free models, as proposed in \citep{Lee&etal:18ICLR,Hendrycks&etal:18ICLR, Ren&etal:19ICLR}.  With simulated OoD samples, it is possible, in principle, to capture fully the difference between the anomaly and anomaly-free distributions and derive  a likelihood ratio test as proposed by \cite{Ren&etal:19ICLR}.  In practice, however, there could be uncountably many OoDs, and simulating OoD samples are highly nontrivial.  A heuristic solution is to perturb training samples from the anomaly-free model and use them to create a proxy of OoD samples.

Existing OoD detection techniques, under the most favorable conditions when training samples are unlimited, learning algorithm most powerful, and the complexity of neural network unbounded, are fundamentally limited in two aspects.  First, in general, there does not exist a uniformly most power test (even asymptotically) for all possible anomaly models. This means that, for every detection rule, there are anomaly cases for which the power of detection is suboptimal. Second, they do not provide Chernoff consistency \citep{Shao:03book} defined as the type I (false positive) and type II (false negative) errors approach to zero as the number of observations increases. For detecting anomalies in times series, Chernoff consistency is essential.   

The source of such apparently fundamental limits arises, perhaps, from the lack of a clear characterization of the OoD model; the standard notion of OoD being something other than the anomaly-free distribution is simply not precise enough to provide a theoretical guarantee. Indeed,  \cite{Lan&Dinh:21} argue that even perfect density models for the anomaly-free data cannot guarantee reliable anomaly detection, and there are ample examples that demonstrate OoD samples can easily fool standard OoD detectors \citep{Goodfellow&Shlen&Szegedy:15}.       To achieve Chernoff consistency or the asymptotic uniformly most power performance, there needs to be a positive ``distance''  between the distributions of the anomaly-free model and those of the anomaly.  It is the constraint on anomaly models being $\epsilon$-distance away in our formulation makes it possible, under ideal training, implementation, and sampling conditions, to achieve Chernoff consistency.  See Sec.~\ref{sec:detection} for one such approach, building on an earlier result of universal anomaly detection \citep{Mestav&Tong:20SPL}.

\section{Innovations Autoencoder}
\label{sec:learning}

\subsection{Parameterization and Dimensionality of Innovations Autoencoder}
\begin{figure}[h]
        \centering
        \scalefig{0.45}\epsfbox{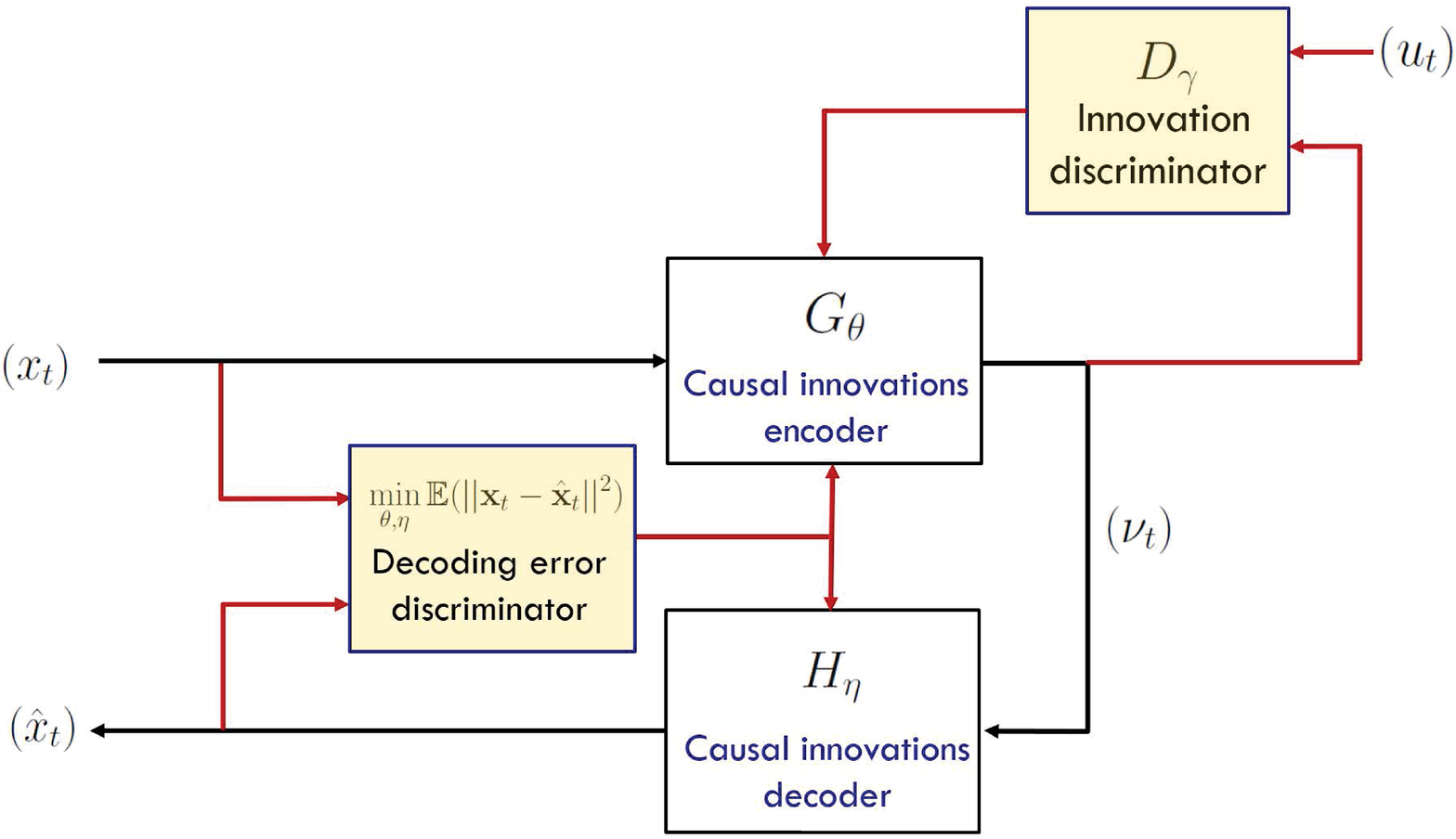}
         \fontsize{8pt}{3pt}
          \input{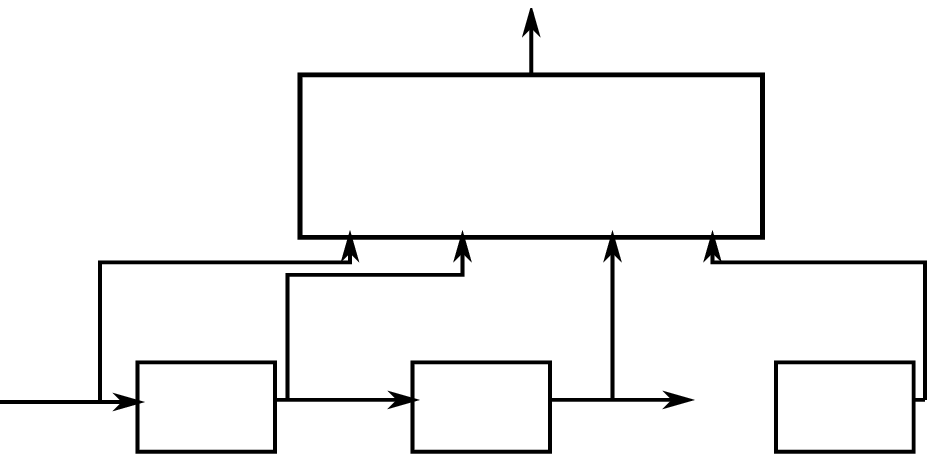}
    \caption{Structure of IAE (left) and a causal CNN implementation (right).}
    \label{fig:Causal}
\end{figure}

 A parameterized  innovations autoencoder (IAE), denoted by $\Ac_{(\theta,\eta)}=(G_\theta, H_\theta)$,  is defined by an {\em innovations encoder} $G_\theta$  and an {\em innovations decoder} $H_\eta$ shown in  Figure~\ref{fig:Causal}(left), both implemented by causal CNNs  \citep{Waibel&etal:89} with parameters $\theta$ and $\eta$ respectively, as shown in Figure~\ref{fig:Causal}(right). Once trained, the innovations sequence $(v_t)$ is produced by the encoder $G_\theta$, and decoded time series $(\hat{x}_t)$  by $H_\eta$.  We note here that the causal encoder and decoder can also be implemented using causal  recurrent neural networks with suitably defined internal states.

 \def\svgwidth{0.5\linewidth}

In a practical implementation, at time $t$, only the  current and a finite number of past samples are used to generate the output.  We define the {\em dimension} of an IAE by the input dimension of its encoder.  For an $m$-dimension IAE $\Ac_{(\theta_m,\eta_m)}$, or $\Ac_m$ in abbreviation, the input of the encoder\footnote{ Note that the index $m$ of $\theta_m$  is associated with the dimension of the autoencoder.  The dimension of $\theta_m$ can be  arbitrarily large. }  $G_{\theta_m}$  is $\xbf^{(m)}_t:=(x_t,\cdots, x_{t-m+1})$ and the output is  a scaler output $\nu_t$ causally produced by $G_{\theta_m}$.  The decoder $H_{\eta_m}$ also takes a finite dimensional input vector $\nubf^{(n(m))}_t = (\nu_t, \cdots, \nu_{t-n(m)+1})$ and causally produces a scaler output $\hat{x}_t$ as an estimate of $x_t$.  Herein, we assume that $n(m) = k_\nu m$ for some design parameter $k_\nu$.

The structure of IAE is similar to some of the existing VAEs in modeling (causally)  sequential data \citep{Bayer&Osendorfer:14arxiv,Chung&etal:15NIPS,Goyal&etal:17NIPS}. The main difference between IAE and these VAEs is the way IAE is trained and the objective of training.  Unlike IAE,  these VAEs aims at obtaining a generative model that does not enforce matching encoder input realizations with those of the decoder output; their objective is to produce the underlying stochastic representations in the forms of probability distributions.

\subsection{Training of IAE: Dimensionality and the Training Objective}
The shaded boxes in Figure~\ref{fig:Causal}(left) represent algorithmic functionalities used in the training process, and the red lines represent input variables from the data flow and output variables used in adapting neural network coefficients.  Two discriminators are used for acquiring the encoder-decoder neural networks.  The {\em innovations discriminator}  is trained via a Wasserstein GAN that evaluates the Wasserstein distance between the estimated innovations $(\nu_t)$ and the standard  (uniform {\it i.i.d.}) sequence.  The {\em decoding error discriminator} evaluates the Euclidian distance between input $(x_t)$ and the decoder output $(\hat{x}_t)$.    The two discriminators generate stochastic gradients in updating the encoder and decoder neural networks.

In learning an $m$--dimensional IAE $\Ac_m$, the two discriminators can only take finite-dimensional samples as their inputs. In practice, the two discriminators may have different dimensions.  For presentation convenience, we assume both discriminators have the sample dimension. We define the {\em dimension of training} as the dimension of the training vectors used by the discriminators to derive updates of the encoder and decoder coefficients.

For an $N$-dimensional training of an $m$-dimensional IAE,  the innovations discriminator $\Dc_{\gamma_m}$  compares  a set of  $N$-dimensional encoder output  samples $\{\nubf^{(N)}_{m,t}:=\Gbf^{(N)}_{\theta_m}(\xbf^{(N+m-1)}_t)\}$   with a set of uniformly distributed i.i.d. samples $\{\ubf_t^{(N)}\}\in[-1,1]^N$ and
produces the empirical gradients of the  Wasserstein distance $W(\nubf^{(N)}_{m,t},\ubf^{(N)}_t)$.  The $N$-dimensional  decoding error discriminator takes decoder outputs $\hat{\xbf}^{(N)}_{m,t}:= \Hbf_{\eta_m}^{(N)}(\nubf_t^{(N+\kappa m-1)})$ and computes the  decoding error $||\hat{\xbf}^{(N)}_{m,t}-\xbf^{(N)}_t||_2$.   The two discriminators compute stochastic gradients and update  encoder, decoder, and discriminator parameters $(\theta_m, \eta_m,\gamma_m)$ jointly. See a pseudocode in Appendix A.

The learning objective of IAE is minimizing  a weighted sum of the Wasserstein distance between the probability distributions of $\nubf^{(N)}_{m,t}$ and $\ubf^{(N)}_t$ and the
decoding error of the autoencoder. By the Kantorovich-Rubinstein Duality, the training algorithm can be derived from the min-max optimization:
\beq \label{eq:IGAN}
\min_{\theta,\eta} \max_{\gamma} \bigg(L_m^{(N)}(\theta,\eta,\gamma) := \mbbE[D_\gamma({\nubf}^{(N)}_{m,t},\ubf^{(N)}_t)]+\lambda \mbbE[ ||\hat{\xbf}^{(N)}_{m,t} - \xbf^{(N)}_t||_2]\bigg),
\eeq
where the first term measure how close the innovation estimated ${\nubf}^{(N)}_{m,t} = \Gbf_{\theta_m}^{(N)}(\xbf^{(N+m-1)}_t )$ is
to a standard reference vector with a uniformly distributed {\it  i.i.d} random vector $\ubf_t^{(N)}$. The second term measures how well   ${\nubf}^{(N)}_{m,t}$ serves as an innovations sequence in reproducing $\xbf^{(N)}_t$.  A  pseudo code that implements the IAE learning is shown in the Appendix.

%
%
%\tcr{Given a set of training samples of $(x_t)$, we replace the ensemble average  with the empirical average based on training samples and solve an empirical risk optimization problem to obtain $(\theta_{m}, \eta_{m}, \gamma_{m})$ for each fixed $m$, where $m$ is the memory size of the networks. The pseudo code that implements such an optimization is shown in Algorithm~\ref{alg:IGAN}.}
%

\subsection{Convergence Analysis}
We will not deal with the convergence of the learning algorithms, which is more or less standard; we shall assume that the learning algorithm converges to its global optimum.  Here we address a ``structural'' convergence issue of some theoretical significance.

A practical implementation of IAE  can only be of a finite dimension $m$. So is the dimension $N$ of the training process.  Such a finite dimensional training can only enforce properties of a finite set of variables of the random process.   Let $\Ac_{m}^{(N)} = (G_{\theta_m^{\tiny (N)}},H_{\theta_m^{\tiny (N)}})$  be  the encoder and decoder output sequences of the $m$-dimensional autoencoder optimally trained with an $N$-dimensional training process  according to (\ref{eq:IGAN}).  Let $\Ac=(G,H)$ be the ideal IAE with encoder $G$  and decoder $H$.  Let $(\nu_t)$ and $(x_t)$ be the output sequences of $G$ and $H$, respectively.   We are interested in how $\Ac_m^{(N)}$ converges to $\Ac$ in some fashion.

 Ideally, we would like to have $\nu_t \xrightarrow{d} \nu_t$ and $x_{m,t}\xrightarrow{L_2} x_t$, which, unfortunately, is not achievable with finite  dimensional training.
 Our goal, therefore, is to achieve a {\em finite-block convergence} defined as follows:

 \begin{Definition}[Finite training-block convergence]
 An $m$-dimensional IAE $\Ac_{m}^{(N)}$  trained with $N$-dimensional training samples converges in training block size $N$ to $\Ac=(G,H)$  if, for all $t$,
\beq
\nubf_{m,t}^{(N)} \xrightarrow{d} \nubf_t^{(N)},~~\xbf_{m,t}^{(N)} \xrightarrow{L_2} \xbf_{t}^{(N)},~~\mbox{as $m\rightarrow \infty$.}
\eeq
\end{Definition}
Note that, even though the dimension $N$ of the learning process can be arbitrarily large, the finite-block convergence does not guarantee the innovations vector of block size greater than $N$ consists of uniform {\it i.i.d.} entries, nor does it ensure that a block of the decoder output of size greater than $N$ can approximate the block of encoder inputs probabilistically unless the process $(x_t)$  has a  short memory.   In practice, unless there is an adaptive procedure to train IAE with increasingly higher dimensions,  one may have to be content with a weaker measure of convergence as defined above.

 Note also that,   by the definition of innovations sequence, it suffices to require that the encoded vector ${\nubf}_{m,t}$ converges in distribution to a vector of uniform {\it i.i.d.} random variables.  A stronger mode of convergence of the decoded sequence is necessary, however.  Herein, we restrict ourselves to the $L_2$  (mean-square)  convergence.

We make the following  assumptions on $\Ac_m^{(N)}$ and $\Ac$:
\ben
\item[A1] {\bf Existence:}  The random process $(x_t)$  has an innovations representation defined in (\ref{eq:G}-\ref{eq:H}), and there exists a causal encoder-decoder pair $(G, H)$ satisfying (\ref{eq:G}-\ref{eq:H}) with $H$ being uniform continuous.
\item[A2] {\bf Feasibility:}    There exists a sequence of finite-dimensional IAE encoding-decoding functions $(G_{\tilde{\theta}_m}, H_{\tilde{\eta}_m})$ that converges uniformly to $(G,H)$ as $m  \rightarrow \infty.$
\item[A3] {\bf Training:} The training sample sizes are  infinite.     The training algorithm for all  finite dimensional  IAE using finite dimensional training samples converges almost surely to the global optimal.
%\item[A4]  {\bf Optimal training:}  The learning model that obtains optimal neural network parameters from the empirical risk minimization of (\ref{eq:IGAN}).
%\item[A5] {\bf Continuity:} Let the two transformation $G_\theta$ and $H_\eta$ be continuous functions for $\forall \theta, \eta$
% \item[A6] {\bf Compactness} Let $x_t$ and $\nu_t$ be taken from compact metrics set respectively, that is $\forall t$, $x_t\in\chi$ and $\nu_t\in V$ where $\chi$ and $V$ are both compact.
\een

With these assumptions, we have the following structural convergence.  See  Appendix A for a proof.
\begin{theorem} \label{thm:converge}
Let $\Ac_{m}^{(N)} = (G_{\theta_m^{\tiny (N)}},H_{\theta_m^{\tiny (N)}})$  be the $m$-dimensional autoencoder optimally trained with training sample dimension $N$ according to (\ref{eq:IGAN}).  Under (A1-A4),  $\Ac_{m}^{(N)}$ converges (in finite block size $N$) to $\Ac$.
\end{theorem}

We now consider the special case of an autoregressive process of finite order $K$ to gain insights into assumptions A1-A4 and Theorem~\ref{thm:converge}.
It is sufficient to demonstrate the case for  the AR(1) process defined by
\[
x_t = \alpha x_{t-1} + \nu_t, ~~\alpha \in (0,1),
\]
where $\nu_t \sim \Uc(-1,1)$ is a uniformly distributed on $[-1,1]$  {\it i.i.d.} sequence. A natural IAE $\Ac=(G_\theta,H_\eta)$ is given by
\begin{align}
G_\theta:~~& \nu_t = G_\theta ( \xbf_t^{(\infty)}) = \thetabf^{\T}  \xbf_t^{(\infty)},~~\thetabf=(1, -\alpha, 0, 0, \cdots), \\
H_\eta:~~ &  x_t = H_\eta( \nubf_t^{(\infty)}) = \etabf^{\T}  \nubf^{(\infty)}_{t},~~\etabf = (1, \alpha, \alpha^2,\cdots).
\end{align}
It is readily verified that both $H$ and $G$ are uniform continuous. Assumption A1 is satisfied.

Now consider the $m$-dimensional IAE  $\tilde{A}_m=(G_{\tilde{\theta}_m},H_{\tilde{\eta}_m})$ defined by
\begin{align}
G_{\tilde{\theta}_m}:~~& \tilde{\nu}_{m,t} = G_{\tilde{\theta}_m} (\xbf_t^{(m)}) = \tilde{\thetabf}_m^{\T}  \xbf_t^{(m)},~~\tilde{\thetabf}_m=(1, -\alpha, 0, 0, \cdots). \\
H_{\tilde{\eta}_m}:~~ &  \tilde{x}_{m,t} = H_\eta(\tilde{\nubf}_{m,t}^{(\kappa_\nu m)}) = \tilde{\etabf}_m^{\T}  \nubf^{(\kappa_\nu m)}_{t},~~\tilde{\etabf}_m = (1, \alpha, \alpha^2,\cdots, \alpha^{\kappa_\nu m-1}).
\end{align}
It is immediate that $G_{\tilde{\theta}_m} \rightarrow G_\theta$ and $H_{\tilde{\eta}_m} \rightarrow H_\eta$ uniformly as $m\rightarrow \infty$.  Therefore assumptions A2-A3 are met.

From (\ref{eq:IGAN}), we have, for all $N$, $m>2$ and $\gamma$,
\[
L_m^{(N)}(\tilde{\theta}_m,\tilde{\eta}_m,\gamma) := \mbbE[D_\gamma(\tilde{\nubf}^{(N)}_{m,t},\ubf^{(N)}_t)]+\lambda \mbbE[ ||\tilde{\xbf}^{(N)}_{m,t} - \xbf^{(N)}_t||_2] =\lambda \mbbE[ ||\tilde{\xbf}^{(N)}_{m,t} - \xbf^{(N)}_t||_2].
\]
Since $H_{\tilde{\eta}_m}$ is the best $l_2$  approximation of $H$, $\mbbE( ||\tilde{\xbf}^{(N)}_{m,t} - \xbf^{(N)}_t||_2)=\min_{\theta,\eta} \mbbE(||\hat{\xbf}^{(N)}_{m,t} - \xbf^{(N)}_t||_2)$. Therefore, $\tilde{A}_m=(G_{\tilde{\theta}_m},H_{\tilde{\eta}_m})$  is a global optimum of (\ref{eq:IGAN}).  Therefore, Theorem~\ref{thm:converge} is verified.  Further, with $\tilde{A}_m$, we have strong convergence of $\tilde{\nu}_{m,t} = \nu_t$ for all $m \ge 2$ and $(\tilde{x}_{m,t}) \xrightarrow{L_2} (x_t)$ as $m\rightarrow \infty$.

\section{Anomalous Sequence Detection via Innovations Autoencoder}
\label{sec:detection}
We develop an IAE-based approach to nonparametric  anomalous sequence detection (or simply anomaly detection for brevity)  and demonstrate in Sec.~\ref{sec:performance} the proposed approach in a smart power grid application  using data collected in a microgrid.

\subsection{A Nonparametric Anomaly Model}
We consider the problem of real-time detection of anomalies in a time series  $(x_t)$ modeled as a random process with unknown temporal dependencies and probability distributions.
Let $\xbf_t$ be a vector consisting of  a finite block of current and past sensor measurements.  Under the null hypothesis $\Hc_0$  that models the anomaly-free measurements and  the alternative $\Hc_1$ for the anomaliies, we consider the following hypothesis testing problem
\beq \label{eq:H0H1}
\Hc_0: \xbf_t \sim f_0~~{\rm\it vs.}~~\Hc_1: \xbf_t \sim f_1 \in \Fmsc_1 = \{f: ||f_1-f_{0}|| \ge \epsilon\}
\eeq
with unknown probability distribution $f_0$ under the anomaly-free hypothesis and a collection $\Fmsc_1$  of unknown anomaly distributions under $\Hc_1$.   Parameter $\epsilon>0$  represents  the degree of separation between the anomaly and anomaly-free models where $\|\cdot\|$ is the total variation distance 
although other measures 
 such as the Shannon-Jensen distance and the Kullback-Leibler divergence are equally applicable.  We assume that an anomaly-free dataset $\Xmsc_0$ is available for offline or online training.

Note that $\Hc_0$ is a simple hypothesis with a single  distribution $f_0$  whereas $\Hc_1$ is a composite hypothesis that captures all possible anomalies in $\Fmsc_1$.  Prescribing a positive distance $\epsilon>0$ between the anomaly-free and the collections of anomaly models is crucial to establish Chernoff consistency that drives the false positive and false negative rates  to zero as the dimension of $\xbf_t$ increases.

\subsection{Anomaly detection via IAE and Uniformity Test}
 A defining feature of IAE is that, under $\Hc_0$, the innovations encoder transforms the measurement time series with an unknown probability model to the standard uniform {\it i.i.d.} sequence.  Through an IAE trained with anomaly-free data, the anomaly detection problem is transformed to testing whether the
transformed sequence is {\it i.i.d.} uniform, for which Chernoff-consistent detectors can be constructed.    Implicitly assumed in this approach is that an anomaly process $\epsilon$-distance away from that of anomaly-free will not be mapped to a uniform {\it i.i.d.} process which is reasonable in practice.

\begin{figure}[h]
\center
\scalefig{0.8}\epsfbox{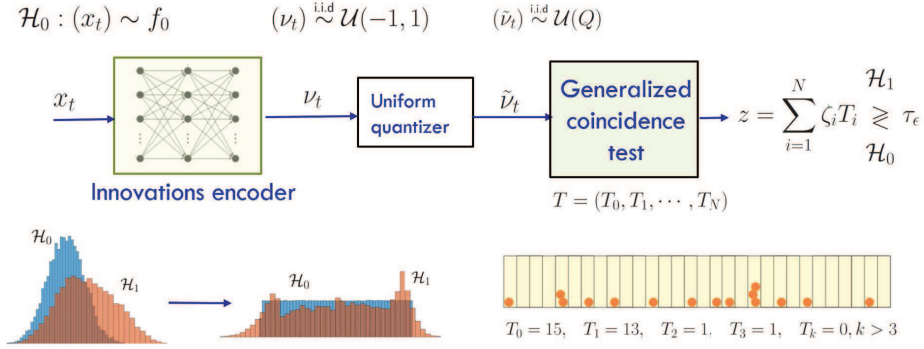}
\caption{\small IAE Uniformity Test. Top: Implementation schematic and test statistics under $\Hc_0$.  Bottom left: Histogram of $(x_t)$ and $(\nu_t)$ under $\Hc_0$ and $\Hc_1$.  Bottom right: an example of coincidence statistics $(T_i)$ with 30 quantization levels and 15 samples.}
\label{fig:uniformity}
\end{figure}

Fig.~\ref{fig:uniformity} shows a schematic of the proposed IAE anomaly detection: the sensor measurements $(x_t)$ is passed through an innovations encoder that, under the anomaly-free hypothesis $\Hc_0$, generates a uniform $\Uc(-1,1)$ {\it i.i.d.}   innovations sequence $(\nu_t)$.  The innovation sequence is then passed through a uniform quantizer that puts $\nu_t$ in one of the $Q$ equal-length quantization bins to produce a sequence of discrete random variables $\tilde{\nu}_t \in \{1,\cdots, Q\}$:
\[
\tilde{\nu}_t = \left\{\begin{array}{ll}
1, & \nu_t \le -1 + 2/Q,\\
i,   &  -1 + 2(i-1)/Q < \nu \le -1+ 2i/Q,~~ i=2, \cdots, Q-1,\\
Q, & \nu \ge 1-2/Q.\\
\end{array}
\right.
\]
Under $\Hc_0$, we thus have a uniform $Q$-ary {\it i.i.d.} sequence $\tilde{\nu}_t$, transforming the original hypothesis testing problem to the following  derived  hypotheses from (\ref{eq:H0H1}):
\beq \label{eq:H0'H1'}
\Hc_0':  \tilde{\nu}_t \sim P_0 = (\frac{1}{Q}, \cdots, \frac{1}{Q})~~\mbox{\rm\it vs.}~~\Hc_1':  \tilde{\nu}_t \sim P_1 \in \Pmsc_1=\{ (p_1, \cdots, p_Q), ||P_1-P_0||_1 \ge \epsilon'\}.
\eeq
where $P_0$ and $P_1$ are $Q$-ary probability mass functions.  Testing $\Hc_0'$ against $\Hc_1'$ is a classic problem \citep{David:50Biometrika,Viktorova:64TPA,Paninski:08TIT,Goldreich:17book}.

Consider (\ref{eq:H0'H1'}) with  $N$ samples $\tilde{\nubf}^{(N)}_t = (\tilde{\nu}_t,\cdots, \tilde{\nu}_{t-N+1})$, a sufficient statistic equivalent to the histogram is the {\em coincidence statistic} $T=(T_0,\cdots, T_N)$ where $T_i$ is the number of quantization bins containing exactly $i$ samples.   See~Fig.~\ref{fig:uniformity} (bottom right) for an example for $Q=30$ and $N=15$.   The coincidence statistics such as $T_0$ and $T_1$ characterize the uniformity property particularly well when samples are ``sparse'' relative to the quantization level.  For instance, when $Q$ is considerably larger than $N$, the $T_1$ value of a uniformly distributed $\tilde{\nu}_t$ tends to be large,  and  $T_0$  tends to be small, relatively.
It is shown in \cite{Paninski:08TIT} that using $T_1$ alone achieves Chernoff consistency, and the sample complexity is roughly at the order of $\sqrt{Q}$.

A general form of a coincidence test is a linear test given by
\beq
\sum_{i=1}^N \zeta_i T_i  \begin{array}{c}\Hc_1\\\gtrless\\\Hc_0\\\end{array} \tau_{\epsilon'},
\label{eq:decision}
\eeq
where the threshold parameter,  a function of $\epsilon'$ (and $N)$, controls the level of false positive rate.  Paninski gives $ \tau_{\epsilon }$  for  the sparse sample case when only $T_1$ is used, whereas  \cite{Viktorova:64TPA} showed the coefficients for the asymptotically most powerful linear detector.

\section{Performance Evaluation}
We present  two sets of evaluations  based on a combination of field collected datasets  from actual systems and synthetic datasets designed to test specific properties.  Sec.~\ref{sec:data} focuses on the IAE encoder-decoder performance using the runs up and down test \citep{Gibbons03}
for the independency of the encoder output sequence and the reconstruction error of the decoded sequence. Sc.~\ref{sec:performance} focuses on the IAE-based anomaly detection in a smart power grid.

\subsection{Training and testing datasets}\label{sec:data}
\begin{figure}[h]
    \centering
    \includegraphics[scale=0.3]{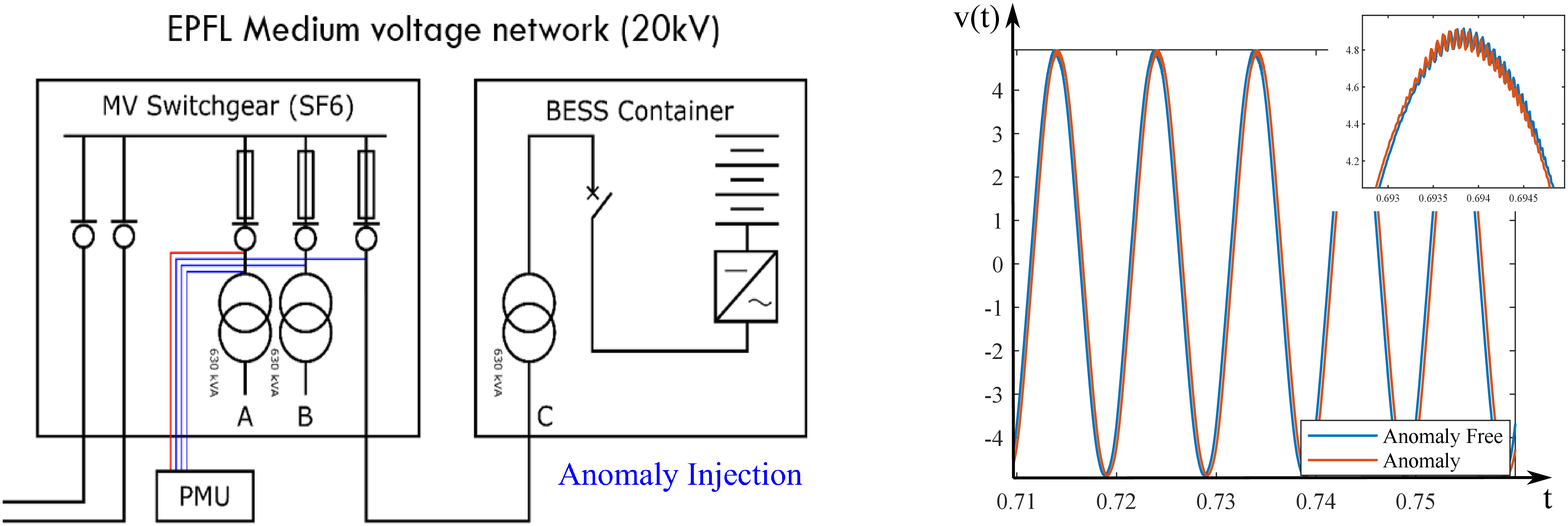}
    \caption{Battery Energy System at EPFL. See \cite{SossanFabrizio&Namor_2016} for detailed description.}
    \label{fig:BESS System}
\end{figure}

We used two field-collected datasets of continuous point-on-wave (CPOW) measurements from two actual power systems.
The BESS dataset contained direct bus voltage measurements sampled at 50 kHz at a  medium-voltage (20kV) substation collected from the
EPFL campus smart grid as described by \cite{SossanFabrizio&Namor_2016}.  As shown in  Fig.~\ref{fig:BESS System} (left),   several circuits  were connected at a bus via a  medium voltage switchgear.  Also connected to the same bus was a battery energy storage system (BESS) used to emulate physically anomaly power injections.  The BESS dataset captured anomaly-free measurements and anomaly power injections that varied from 0 to 500 (kW).
 Fig.~\ref{fig:BESS System} (right) shows segments of anomaly and anomaly-free measurements of a single phase CPOW voltage waveforms.  The CPOW samples exhibited narrow-band (sinusoidal-like) characteristics with strong temporal correlations.
 Because of the frequency and voltage regulation mechanisms in a power system, the voltage
magnitudes and frequencies were tightly controlled such that both anomaly-free and anomaly voltage CPOW data were very similar although a zoomed-in plots exhibited differences in high-order harmonics, as shown in the zoom-in plot of Fig.~3 (right). The detection of anomaly in such voltage CPOW measurements was quite challenging.

% \vskip -0.in

The second field-collected dataset (UTK)  contained direct samples of voltage waveform at 6 kHz collected at the University of Tennessee.  Only anomaly-free CPOW measurements were available.  Similar to the BESS dataset, the UTK dataset contained strongly correlated narrow-band  samples.

Besides the two field datasets (BESS and UTK), we also designed several synthetic datasets to evaluate specific properties of IAE and IAE-based anomaly detections.  These datasets are described in Sec.~\ref{sec:performance} and Sec.~\ref{sec:Anomaly}.

\subsection{IAE performance}\label{sec:performance}
We evaluated the performance of IAE and several of benchmarks for extracting innovations sequences.  In particular, we examined whether the estimated innovation sequences passed the test of being statistically independent and identically distributed.  We also evaluated the mean-squared error (MSE) of the
reconstructed signal.

\subsubsection{Benchmarks}
Since there were very few  techniques specifically designed for extracting innovation sequences,  we compared IAE with four benchmarks adapted from existing techniques aimed at extracting independent components.
Among these benchmarks, three were deep learning solutions (NLLS, ANICA, f-AnoGAN) and two of those (ANICA, f-AnoGAN) were autoencoder based.
\bitem
\item {\em LLS:} LLS was a linear least-squares  prediction error filter that generated the one-step prediction error sequence.  For stationary Gaussian time series, a perfectly trained LLS predictor would produce a true innovations sequence.
\item {\em NLLS:} NLLS was a nonlinear least-squares prediction error filter that generated the one-step production error time series.   If the measurement time series was obtained from a sampled (possibly) non-Gaussian process with additive Gaussian noise, the NLLS prediction error sequence would be a good approximation of an innovations process.
\item {\em ANICA:}  ANICA was an adaption of the nonlinear ICA autoencoder proposed in \citep{Brakel&Bengio:17}.  Aimed to extract independent components from a block of measurements, the original design did not enforce causality and was not intended to generate an innovations sequence.
\item {\em f-AnoGAN:}  f-AnoGAN proposed by \cite{Schlegl&Seebock:19} was an autoencoder technique involving convolutional neural networks.  The goal was to extract low-dimensional latent variables as features from which the decoder could recover the original. Since the autoencoder was trained with anomaly-free samples, the intuition was that anomaly data would have high reconstruction errors. Such a construction could be viewed as a nonlinear principle component analysis (PCA).
\eitem

IAE was implemented by adapting the Wasserstein GAN with a few modifications \footnote{\url{https://keras.io/examples/generative/wgan_gp/}}. For all cases in this paper, similar neural network structures were used: the encoder and decoder both contained three hidden layers with 100, 50, 25 neurons respectively with hyperbolic tangent activation. The discriminator contained three hidden layers with 100, 50, and 25 neurons, of which the first two used hyperbolic tangent activation and the last one the linear activation. The tunning parameter used for each case is presented in the Appendix.

\begin{table}[h]
\begin{center}
\begin{small}
\begin{sc}
\begin{tabular}{ll}
\toprule
Dataset & Model \\
\midrule
Moving Average (MA):    &$x_t=\frac{1}{10}\sum_{i=1}^{10} \nu_{t-i}$  \\
Linear Autoregressive (LAR) &$x_t=0.5 x_{t-1}+\nu_t$ \\
Nonlinear Autoregressive  (NLAR)    &$x_t=0.5 x_{t-1}+0.4\mathbbm{1}(x_{t-2}<0.7) +\nu_t$  \\
\bottomrule
\end{tabular}
\end{sc}
\caption{Test Synthetic Datasets. $\nu_t\stackrel{\tiny\rm i.i.d}{\sim}\mathcal{U}[0,1]$. $\mathbbm{1}(\cdot)$ is the indicator function.}
\label{tb:Synthetic dataset}
\end{small}
\end{center}

\end{table}

\subsubsection{Test datasets}
Besides the two field datasets (BESS and UTK) described in Sec.~\ref{sec:data}, we included three synthetic datasets shown in Table.~\ref{tb:Synthetic dataset} to produce different levels of temporal dependencies and probability distributions in test data.
 In particular, the linear autoregressive (LAR) dataset was chosen such that a properly trained LLS approach
 would produce an innovations sequence. For the moving average (MA) and nonlinear autoregressive (NLAR) datasets, sufficiently complex
 neural network implementation of NLLS and ANICA could produce approximations of the innovations sequences. 
 
  For all the synthetic cases, we used the memory size m = 20 dimensional IAE and training sample dimension n = 60 in the neural network training, and 100,000 samples were used for training for all cases.
The neural network memory size for real data cases were chosen to be m = 100, n = 250, due to stronger temporal dependency.

\subsubsection{Performance and discussion}
In evaluating the performance of the benchmarks, we adopted  the runs up and down test that used the numbers of consecutively ascending or descending samples as test statistics of statistical independence.
It was shown (empirically) to have the best performance in \citep{Gibbons03}.  We also evaluated the empirical  mean-squared error of the reconstruction.

\begin{table}[h]
\begin{center}
\begin{small}
\begin{sc}
\begin{tabular}{lccccc}
\toprule
Method  &MA &LAR &NLAR  &UTK &BESS \\
\midrule
IAE (p-value)         &0.9492     &0.8837     &0.7498     &0.9990 &0.8757\\
LLS (p-value)       &0.3222     &0.9697     &0.0186     &$<$0.0001  &$<$0.0001\\
NLLS  (p-value)       &0.2674     &N/A        &0.5116   &$<$0.0001   &$<$0.0001\\
ANICA  (p-value)     &$<$0.0001  &0.1502  &$<$0.00019   &$<$0.0001   &$<$0.0001\\
F-anoGAN  (p-value)   &$<$0.001   &0.0106 &$<$0.0001  &$<$0.0001  &$<$0.0001 \\\midrule
IAE(MSE) &6.3849 &8.5366 &9.398425   &14.5641    &21.0144\\
Anica(MSE)   &137.2839   &274.3765   &283.31250 &315.6521   &319.9284\\
F-anoGAN(MSE) &6.7421 &12.4379 &11.6458 &11.8630 &11.8821\\
\bottomrule
\end{tabular}
\end{sc}
\caption{p-value of the runs test and the mean-squared error (MSE) of the reconstruction.}
\label{tb:p-value}
\end{small}
\end{center}
\vskip -0.1in
\end{table}

Table.~\ref{tb:p-value} shows the p-values of the runs up and down test. NLLS prediction method was not implemented for LAR case because the linear least-square was sufficient for demonstration purposes. As autoencoder based methods, F-anoGAN and IAE achieved the comparable reconstruction error, with F-anoGAN performing better. ANICA failed to obtain a competitive reconstruction error. 

As for the independence test for the BESS dataset, IAE achieved the highest p-values for all the scenarios except the synthetic LAR dataset designed specifically for the LLS algorithm. For the synthetic datasets, LLS and NLLS produced sequences that the runs tests could not easily reject the independence hypothesis. For the field datasets, LLS and NLLS failed the run tests. Not specifically designed for extracting innovations, ANICA failed the run tests for statistical independence.

\subsection{Detection of anomalies in power systems}\label{sec:Anomaly}
We evaluated the performances of several benchmarks in detecting system anomalies in field-collected dataset BESS, the UTK dataset with synthetically generated anomalies, and two synthetic time series datasets. We compared benchmark techniques using
their receiver characteristic curves (ROC) that plotted true positive rates (TPR) over a range of the false positive rate (FPR). The area under ROC (AUROC) was also calculated for all techniques.

% \vskip -0.in
\subsubsection{Test datasets}
In addition to the BESS dataset that included both anomaly and anomaly-free measurements, we also considered three additional datasets shown in Table~\ref{tb:Synthetic Test}, two synthetic datasets ({\tt SYN1, SYN2}) and one semi-synthetic dataset with anomaly waveforms added to the field-collected anomaly-free samples \citep{Wang&Liu&Tong:21TPS}.     {\tt SYN1} and {\tt SNY2} had the identical anomaly-free models of AR(1) Gaussian.  Under the anomaly hypothesis, {\tt SYN1} was AR(2) Gaussian, whereas {\tt SYN2} was an AR(1) with uniform innovations.  Because only the anomaly-free training samples were assumed and the anomaly waveforms and probability distributions were arbitrary, the same anomaly detector trained based on the anomaly-free samples were tested under {\tt SYN1} and {\tt SYN2}.

\begin{table}[h]
\begin{center}
\begin{small}
\begin{sc}
\begin{tabular}{lccl}
\toprule
Test Case & Anomaly Free Samples & Anomaly Samples &Block Size ($N$) \\
\midrule
Syn1    &$x_t=0.5 x_{t-1}+\nu_t$    &$x_t=0.3x_{t-1}+0.3x_{t-2}+\nu_t$  &1000 \\
Syn2    &$x_t=0.5 x_{t-1}+\nu_t$    &$x_t=0.5 x_{t-1}+\nu'_t$ &1000\\
UTK     &Real Data                  &GMM Noise                  &200\\
BESS    &Real Data                  &Real Data                  &500\\
\bottomrule
\end{tabular}
\end{sc}
\caption{Data Detection Test Cases. $\nu_t\stackrel{i.i.d}{\sim}\mathcal{N}(0,1)$,~~$\nu'_t\stackrel{i.i.d}{\sim}\mathcal{U}[-1.5,1.5]$}
\label{tb:Synthetic Test}
\end{small}
\end{center}
%\vskip -0.1in
\end{table}

\subsubsection{Benchmarks}
Few benchmark techniques were suitable for the anomaly sequence
detection problem considered here. Most relevant prior techniques that could be applied directly were the one-class support vector machine (OCSVM) proposed in \citep{Scholkopf:99NIPS} and f-AnoGAN  in \citep{Schlegl&Seebock:19}.  OCSVM, a semisupervised classification technique, was implemented with radial basis functions as its kernel and was trained with anomaly-free samples.  Although not designed as an anomaly detection solution,  ANICA \citep{Brakel&Bengio:17}  was adapted to be a preprocessing algorithm (similar to IAE) before applying a uniformity test described in Sec.~\ref{sec:detection}.

We have also included the Quenouille test \citep{Priestley:81Book} designed to test the goodness of fit of an AR($k$) model (with {\tt SYN1} dataset.)  Because of the  asymptotic equivalence of the Quenouille test and the maximum likelihood test of \cite{Whittle51:thesis}, we used Quenouille test as a way to calibrate how well IAE and other nonparametric  tests would perform under AR($k$) time series models with dataset {\tt SYN1}  for which the Quenoulle test is asymptotically optimal.

\subsubsection{Performance on the BESS dataset}
\begin{figure}[h]
    \centering
\includegraphics[scale=0.8]{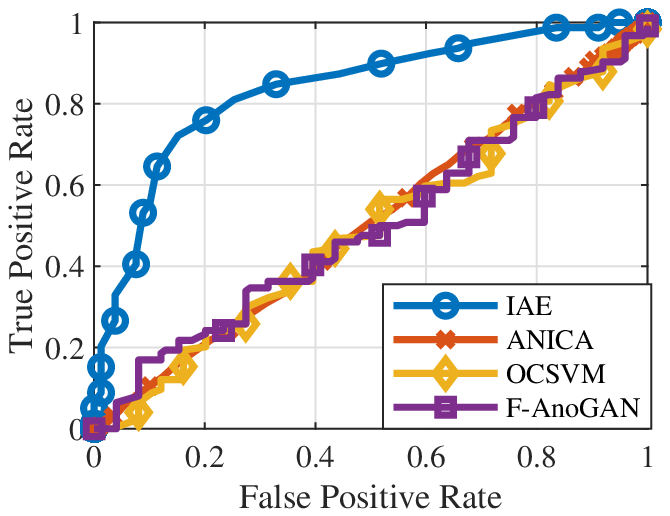}
\includegraphics[scale=0.55]{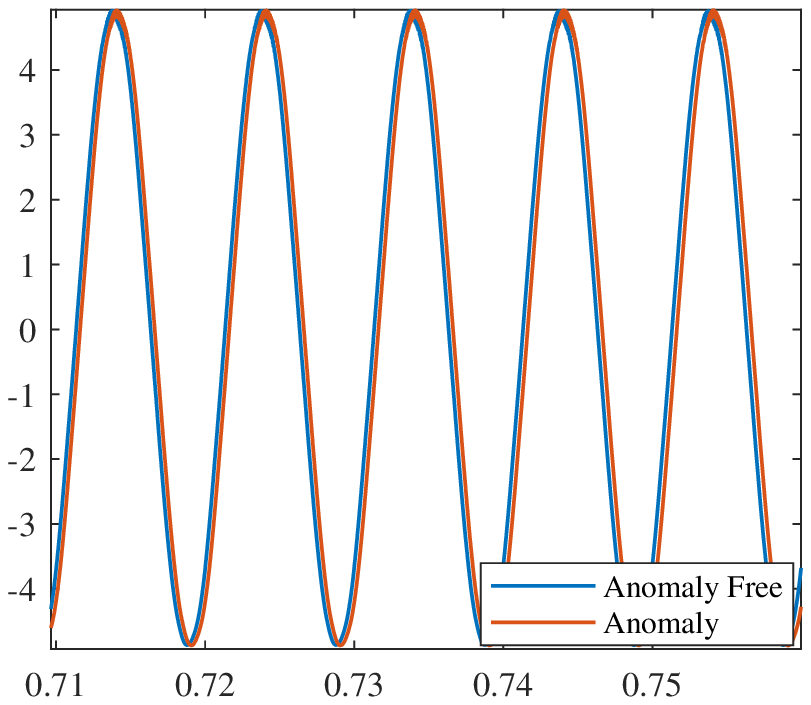}
    \caption{\small Detection performance  for the BESS dataset. Left: ROC curves. (AUROC: IAE:0.8354, ANICA:0.5027 OCSVM:0.4903 F-AnoGAN:0.4993)   Right: Anomaly and anomaly-free traces.}
    \label{fig:Real-BESS}
\end{figure}

The BESS dataset was used to test the proposed testing technique's ability to detect system anomalies. As shown in Fig.~\ref{fig:Real-BESS}, the anomaly and anomaly-free voltage signals were very similar due to the voltage regulation of the bus voltage in the power system. The detection based solely on the raw voltage signal can be very challenging. Fig.~\ref{fig:Real-BESS} (right)  shows the ROC curves obtained using 500-sample blocks. Since the anomaly and anomaly-free samples are hard to distinguish, all the other methods apart from IAE didn't seem to work well in this case.

\subsubsection{Performance on the UTK dataset}
\begin{figure}[h]
    \centering
\includegraphics[scale=0.8]{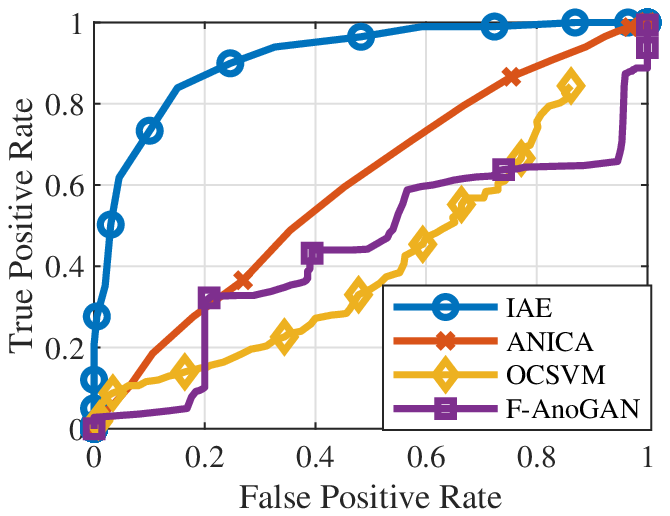}
\includegraphics[scale=0.4]{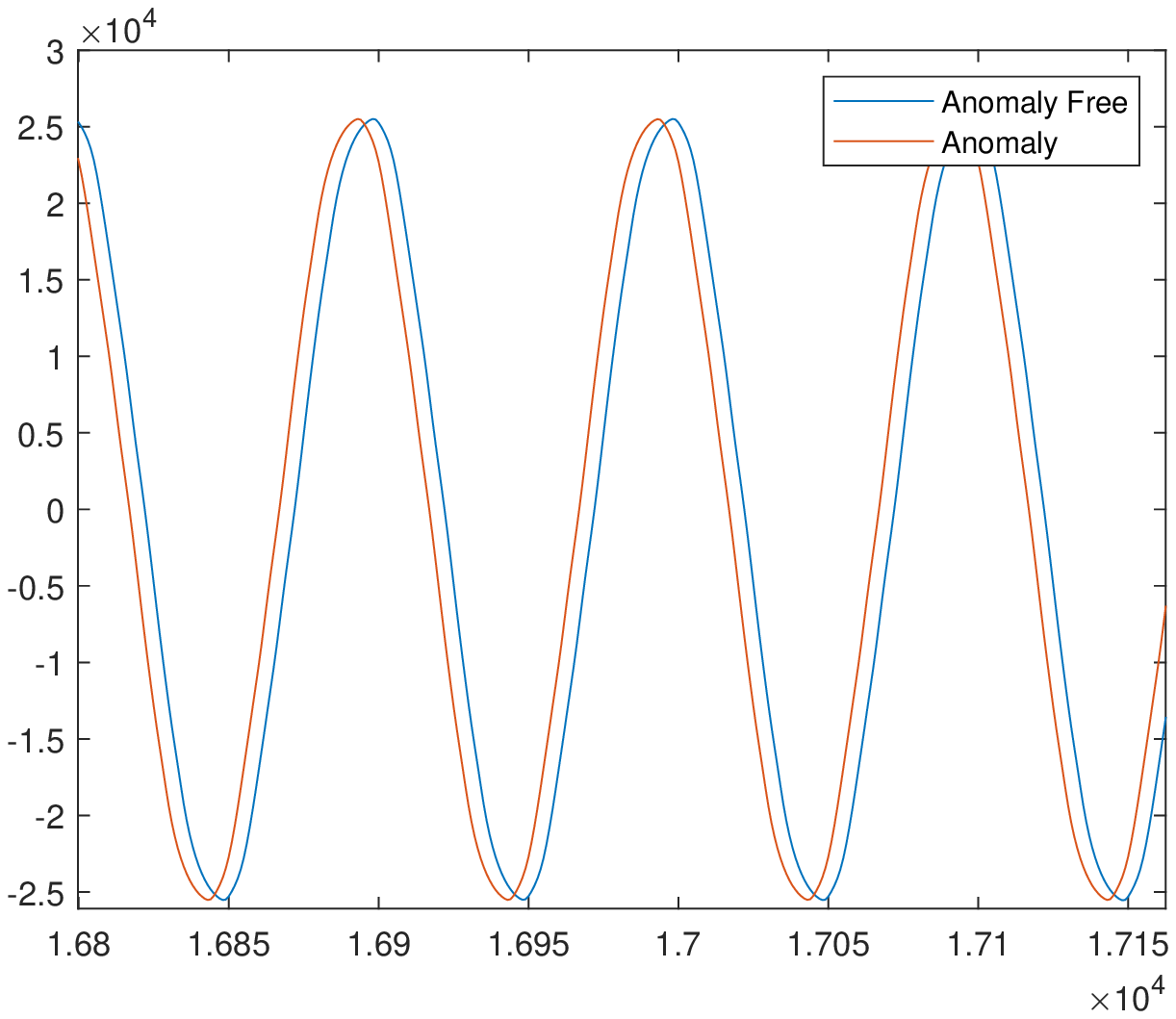}
    \caption{\small Detection performance  for the UTK dataset.  Left: ROC curves. (AUROC: IAE:0.9135, ANICA:0.5967, OCSMV:0.2978, F-AnoGAN:0.4393)   Right: Anomaly and anomaly-free traces.}
    \label{fig:Real-UTK}
\end{figure}

We evaluated benchmark performance on the UTK dataset with synthetic anomaly test samples. To construct the anomaly samples, we added a comparably small Gaussian Mixture noise on the anomaly-free measurements. The signal to noise ratio of the Gaussian Mixture noise to the anomaly free signal is roughly 40dB, and the time-domain trajectories of anomaly and anomaly-free signals are shown in Fig.~\ref{fig:Real-UTK} (right), which demonstrate the level of similarity between anomaly and anomaly-free samples. Seen from Fig.~\ref{fig:Real-UTK} (left), IAE was the only detection method that was able to make reliable decisions, with ANICA performing slightly better than the rest.

%
%   \begin{figure}[h]
%   \centering
%    \begin{subfigure}[t]{0.45\linewidth}
%        \centering
%        \includegraphics[scale=0.6]{figs/ROC-UTK.eps}
%        \caption{ROC Curves of UTK Dataset}
%        \label{fig:ROC-UTK}
%    \end{subfigure}
%        \begin{subfigure}[t]{0.45\linewidth}
%        \centering
%        \includegraphics[scale=0.3]{figs/UTK-t.eps}
%        \caption{Time Domain Trajectory of UTK Dataset}
%        \label{fig:t-UTK}
%    \end{subfigure}
%        \label{fig:Real-UTK}
%        \caption{Anomaly Detection for UTK Dataset}
%    \end{figure}

\subsubsection{Performance on synthetic datasets {\tt SYN 1} and {\tt SYN 2}}
We also conducted anomaly detection based on synthetic data generated by auto-regressive models ({\tt SYN1} and {\tt SYN2}). For both {\tt SYN1} and {\tt SYN2}, the anomaly free datasets were the same, and the two anomaly datasets were designed to highlight the the performance of the detectors facing different anomalies.

% We designed two synthetic data cases with both anomaly samples and anomaly free samples generated from linear auto-regressive processes. In test cases SYN1, $\forall t$, $x_t\sim \mathcal{N}(0,\frac{4}{3})$ for both anomaly and anomaly free data. Since they have the same marginal distribution, testing technique based on one single observation won't work for this case. In the second synthetic test case, anomaly and anomaly free data have the same auto-correlation, thus any spectrum based method won't work well for SYN2. To establish a comparable benchmark for the synthetic cases, we adopt the Quenouille's goodness of fit test (hereafter referred to as Quenouille) which is specifically designed for AR processes. The goodness of fit test proposed by Quenouille used a linear combination of sample auto-correlation as the test statistics $Q$, and under the assumption of anomaly free data being generated from a $k$th order auto-regressive process, $Q$ follows chi-square distribution.

% Another dataset consisting of CPOW measurements is obtained from a Smart Grid with Battery Energy Storage System (BESS) connected in EPFL. The dataset contains 50Hz voltage signals sampled at 50KHz, which exhibits strong temporal dependency. The dataset contains a set of normal data with the battery idle and several scenarios of anomaly measurements where the battery system suddenly inject active or reactive power into the system.

\begin{figure}[ht]
\centering
\includegraphics[scale=0.8]{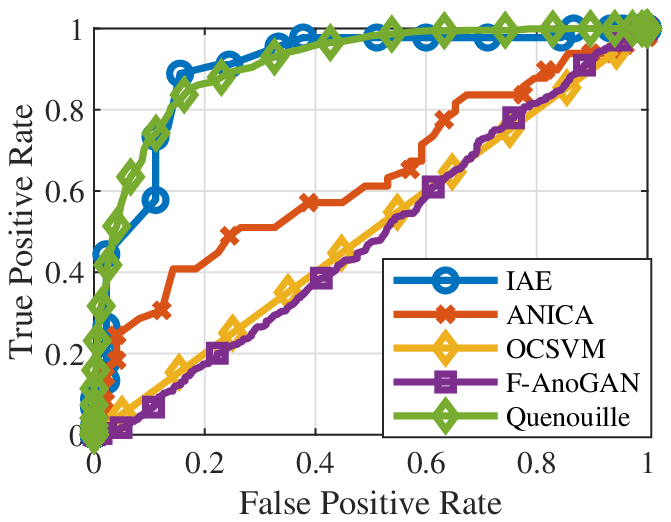}
\includegraphics[scale=0.8]{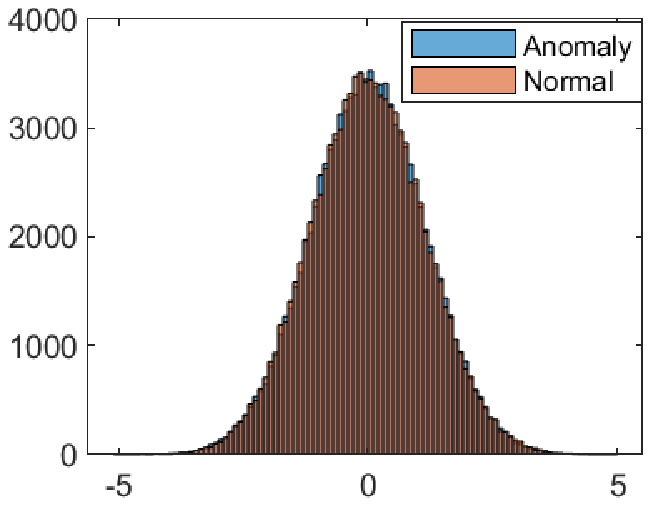}
    \caption{Anomaly Detection for {\tt SYN1}.  Left: ROC curves. (AUROC: IAE:0.9021, ANICA:0.6337, OCSVM:0.4455, F-AnoGAN:0.4881, Quenouille:0.9112)   Right: Anomaly and anomaly-free histograms.}
      \label{fig:Syn1}
 \end{figure}

 In {\tt SYN1},  we designed the anomaly to have the same marginal distribution as the anomaly-free data ($x_t\sim\mathcal{N}(0,4/3)$), as shown in Fig.~\ref{fig:Syn1} (right), intentionally making detection based on a single sample ineffective. As shown by Fig.~\ref{fig:Syn1} (left), IAE performed similarly well as the asymptotically optimal detector (Quenouille). ANICA performed better (with AUROC above 0.5) than the other two machine learning-based detection methods. Because the marginal distributions of the measurements are the same for both hypotheses, the other two machine learning-based techniques were not competitive under this setting.

%
%
%    \begin{subfigure}[t]{0.45\linewidth}
%    \centering
%        \includegraphics[scale=0.6
%        ]{figs/ROC-Syn1.eps}
%        \caption{{\tt SYN1} ROC Curves}
%        \label{fig:ROC-Syn1}
%    \end{subfigure}
%    \begin{subfigure}[t]{0.45\linewidth}
%    \centering
%        \includegraphics[scale=0.6]{figs/AR_hist_syn1.eps}
%        \caption{{\tt SYN1} histograms}
%        \label{fig:hist-Syn1}
%    \end{subfigure}
%    \label{fig:Syn1}
%    \caption{Anomaly Detection for {\tt SYN1}}
%\end{figure}
%
%\begin{figure}
%    \centering
%        \begin{subfigure}[t]{0.45\linewidth}
%    \centering
%        \includegraphics[scale=0.6
%        ]{figs/ROC-Syn2.eps}
%        \caption{{\tt SYN2} ROC Curves}
%        \label{fig:ROC-Syn2}
%    \end{subfigure}
%    \begin{subfigure}[t]{0.45\linewidth}
%    \centering
%        \includegraphics[scale=0.6]{figs/AR_hist_syn2.eps}
%        \caption{{\tt SYN2} histograms}
%        \label{fig:hist-Syn2}
%    \end{subfigure}
%    \caption{Anomaly Detection for {\tt SYN1}}
%    \label{fig:Syn2}
%\end{figure}

\begin{figure}[ht]
\centering
\includegraphics[scale=0.8]{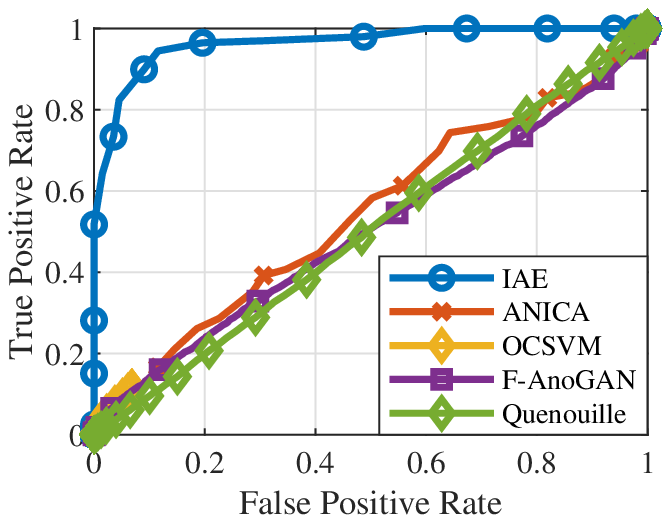}
\includegraphics[scale=0.8]{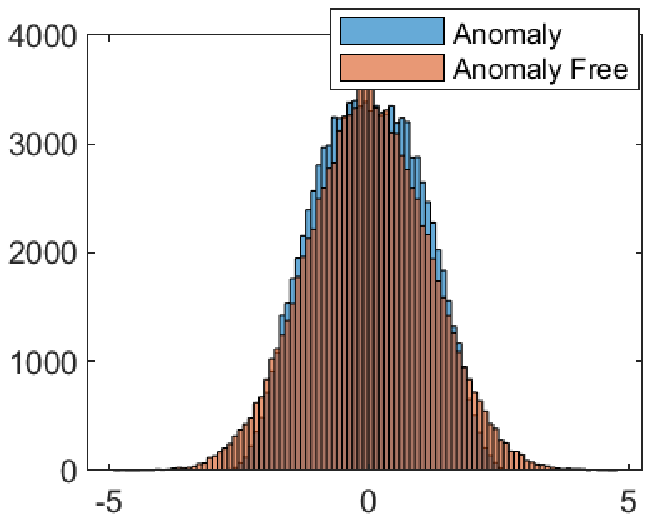}
    \caption{Anomaly Detection for {\tt SYN1}.  Left: ROC curves. (AUROC: IAE:0.9635, ANICA:0.6337, OCSVM:0.5407, F-AnoGAN:0.5020, Quenouille:0.5026)   Right: Anomaly and anomaly-free histograms.}
      \label{fig:Syn2}
 \end{figure}

{\tt SYN2} adopted two auto-regressive models with the same parameters in temporal dependencies.  The marginal distributions of the measurements, however, were slightly different under the anomaly and anomaly-free models. See. Fig~\ref{fig:Syn2} (right), which made it very challenging for OCSVM and F-AnoGAN.  ANICA was also not effective in extracting independent components,  causing failures in the uniformity test. Only IAE was able to capture the difference between the two datasets through the extraction of innovations and made reasonably reliable decisions, as shown in Fig~\ref{fig:Syn2} (left).

\section{Conclusion}
\label{sec:conclusion}
IAE is a machine learning technique that extracts innovations sequence from real-time observations. When properly trained, IAE can serve as a font-end processing unit that transforms processes of unknown temporal dependency and probability structures to a standard uniform {\it i.i.d} sequence.  (Extension to other marginal distributions is trivial.)  IAE is, in someway, an attempt to  realize Wiener's original vision of encoding stationary random processes with the simplest possible form, although  the existence of such an autoencoder is not guaranteed \citep{Rosenblatt:59}.  From an engineering perspective, however, the success of Wiener and Kalman filtering in practice is a powerful  testament that many applications in practice can be approximated by innovations representations.  It is under such an assumption that IAE serves to remove the modeling assumptions in Wiener and Kalman filtering and pursues a data-driven machine learning solution. As an example, the IAE-based anomaly detection is shown to be quite effective for the one-class anaomalous time series detection problem, for which there are few solutions.

\acks{We would like to acknowledge support for this project
from the National Science Foundation (NSF grant 1932501 and 1816397). }

\newpage

% Acknowledgements should go at the end, before appendices and references

% \acks{We would like to acknowledge support for this project
% from the National Science Foundation (NSF grant IIS-9988642)
% and the Multidisciplinary Research Program of the Department
% of Defense (MURI N00014-00-1-0637). }


\vskip 0.2in
\bibliography{BIB_ICAGAN}

\newpage

\appendix
\section{Proof of Theorem 1}
\label{app:theorem}
{\em Proof:}    By A2, there exists a sequence of IAEs $\tilde{\Ac}_m=(G_{\tilde{\theta}_m}, H_{\tilde{\eta}_m})$ of dimension $m$  that converges to $\Ac$.   Let $(\tilde{\nu}_{m,t})$ be the output sequences of $G_{\tilde{\theta}_m}$ and $(\tilde{x}_{m,t})$  the output sequence of the decoder $H_{\eta_m}$.   Similarly defined are vector outputs of the encoder and decoder $\tilde{\xbf}_{m,t}^{(N)}$ and $\tilde{\nubf}_{m,t}^{(N)}$.

We prove Theorem 1 in two steps:
\ben
\item {\bf Finite-block convergence of $\tilde{\Ac}_m$:}   By assumption A2, the uniform convergence of
$G_{\tilde{\theta}_m} \rightarrow G$ implies that, for every $\epsilon_1>0$, there exists   $M_1\in\mathbb{N}^+$  such that,  for all realizations $\xbf^{(\infty)}_t$,  $m>M_1$ and $t \in \mathbb{N}$,
\[
|\tilde{\nu}_{m,t}-\nu_t|< \epsilon_1~~\Rightarrow~~  \mbbE\left(\|\tilde{\nubf}_{m,t}^{(N)}-\nubf^{(N)}_t\|^2\right) \le N\epsilon_1.
\]
 Therefore, for all finite $N$,  we have the following uniform convergence as $m\rightarrow \infty$:
 \beq \label{eq:nutilde}
 \tilde{\nubf}_{m,t}^{(N)} \xrightarrow{L_2} \nubf_t^{(N)}~~\Rightarrow~~  \tilde{\nubf}_{m,t}^{(N)} \xrightarrow{d} \nubf_t^{(N)},~~\Rightarrow~~W(\tilde{\nubf}_{m,t}^{(N)},\nubf_t^{(N)}) \rightarrow 0.
 \eeq

Next we consider the decoder convergence.  Fix $\epsilon_2>0.$
\begin{align*}
|x_t-\tilde{x}_{m,t}| &= |H\circ G(\xbf_t^{(\infty)}) - H_{\tilde{\eta}_m} \circ G_{\tilde{\theta}_m}(\xbf_t^{(m)})|\\
&\le |H\circ G(\xbf_t^{(\infty)}) - H \circ G_{\tilde{\theta}_m}(\xbf_t^{(m)})|\ + |H \circ G_{\tilde{\theta}_m}(\xbf_t^{(m)}) - H_{\tilde{\eta}_m} \circ G_{\tilde{\theta}_m}(\xbf_t^{(m)})|.
\end{align*}
Because $H$ is uniform continuous, there exists an $M_2(\epsilon_2)$ such that for all $m > M_2(\epsilon_2)$ and $\xbf_t^{(\infty)}$,
\[
|H\circ G(\xbf_t^{(\infty)}) - H \circ G_{\tilde{\theta}_m}(\xbf_t^{(m)})|\  < \epsilon_2/2.
\]
 Because $H_{\eta_m}$ converges to $H$ uniformly, there exists an $M_2'(\epsilon_2)$ such that
 \[
 |H \circ G_{\tilde{\theta}_m}(\xbf_t^{(m)}) - H_{\tilde{\eta}_m} \circ G_{\tilde{\theta}_m}(\xbf_t^{(m)})| < \epsilon_2/2,
 \]
 for all $m > M_2'(\epsilon_2)$ and $\xbf_t^{(\infty)}$.  Therefore,  for all $m > \max\{ M_2(\epsilon_2),M_2'(\epsilon_2)\}$,
\[
|\tilde{x}_{m,t}-x_t|< \epsilon_2~~\Rightarrow~~  \mbbE\left(\|\tilde{\xbf}_{m,t}^{(N)}-\xbf^{(N)}_t\|^2\right) \le N\epsilon_2~~\Rightarrow~~ \tilde{\xbf}_{m,t}^{(N)} \underset{m\rightarrow \infty}{\xrightarrow{L_2}} \xbf^{(N)}_t.
\]
The risk $\tilde{L}_m^{(N)} $  converges uniformly:
\beq \label{eq:Ltilde}
\tilde{L}_m^{(N)} :=  W(\tilde{\nubf}_{m,t}^{(N)},\nubf_t^{(N)}) +  \mbbE(\|\tilde{\xbf}^{(N)}_{m,t}-\xbf_t^{(N)}\|_2^2) \xrightarrow{m \rightarrow \infty} 0.
\eeq
%&= \sup_{\gamma} \bigg(\mbbE(D_\gamma (\tilde{\nubf}_{m,t}^{(N)})) - \mbbE(D_\gamma (\nubf_t^{(N)}))\bigg) + \mbbE(\|\tilde{\xbf}^{(N)}_{m,t}-\xbf_t^{(N)}\|_2^2) \nn\end{align}

\item {\bf  Finite-block convergence of $\Ac_m^{(N)}$:}  Fix the dimension of training at $N$.  From the finite-block convergence of $\tilde{\Ac}_m$, $\forall \epsilon$,   there exists  $M_\epsilon$ such that, for all $m> M_\epsilon$,
\beq
 \tilde{L}_m^{(N)} = W(\tilde{\nubf}^{(N)}_{m,t}-\nubf_t^{(N)})+ \mbbE\left(\|\tilde{\xbf}_{m,t}^{(N)}-\xbf^{(N)}_t\|_2^2\right) \leq \epsilon.
\eeq
By the Kantorovich-Rubinstein duality theorem,
\[
W(\tilde{\nubf}^{(N)}_{m,t},\nubf_t^{(N)})
= \max_\gamma \mbbE(D_\gamma (\tilde{\nubf}_{m,t}^{(N)}, \nubf_{t}^{(N)})).
\]
From (\ref{eq:IGAN}),   let (without loss of generality assuming $\lambda=1$)
\[
L_m^{(N)} :=\min_{\theta,\eta}\max_\gamma \bigg(
\mbbE(D_\gamma (\tilde{\xbf}_{m,t}^{(N)},\nubf_{t}^{(N)})
+  \mbbE(\|\xbf^{(N)}_{m,t}-\xbf_t^{(N)}\|_2^2)\bigg). \nn
\]
We therefore have, for all $m \ge M_\epsilon$,
\[
 L_m^{(N)} \le \tilde{L}_m^{(N)} \le \epsilon~~\Rightarrow~~
 \left\{\begin{array}{l}
 W(\nubf_{m,t}^{(N)}, \nubf_t^{(N)}) \le \epsilon,\\
 \mbbE(\|\xbf^{(N)}_{m,t}-\xbf_t^{(N)}\|^2_2) \le \epsilon,
 \end{array}
 \right.
\]
which completes the proof.  \hfill $\Box\Box\Box$

\een

\section{Pseudocode}\label{sec:code}
\begin{algorithm}[h]
   \caption{Training the Innovations Autoencoder}
   \label{alg:IGAN}
\begin{algorithmic}
   \STATE {\bfseries Input:} data $(x_t)$, encoder $H_\eta$, generator $G_\theta$, discriminator $D_\gamma$. $\lambda$ is the gradient penalty coefficient, $\mu$ the weight for auto-encoder, and $\alpha,\beta_1,\beta_2$ hyper-parameters for Adam optimizer.
   \WHILE{Not converged}
  \FOR{$t=1,\cdots,n_{c}$}
        \FOR{$i=1,\cdots,B$}
                \STATE Sample $\xbf_i$ from the input matrix $(x_t)$
                \STATE Sample \textbf{$\mathbf{u}$}$=[u_1,\cdots,u_{n-m+1}]^T\stackrel{i.i.d}{\sim}\mathcal{U}[-1,1]$
                \STATE Sample $\epsilon\sim\mathcal{U}[0,1]$
                \STATE $\hat{\nubf}\leftarrow \mathbf{G}_\theta(\mathbf{x}_i)$
                \STATE $\Bar{\nubf}\leftarrow \epsilon \mathbf{u}+(1-\epsilon)\hat{\nubf}$
                \STATE $L^{(i)}\leftarrow \mathbf{D}_\gamma(\hat{\nubf})-\mathbf{D}_\gamma(\mathbf{u})+\lambda(\lVert \nabla_{\gamma} \mathbf{D}_{\gamma}(\Bar{\nubf})\rVert_2-1)^2$
        \ENDFOR
    \STATE $\gamma\leftarrow Adam(\nabla_\gamma\frac{1}{B}\sum_{i=1}^B L^{(i)},\alpha,\beta_1,\beta_2)$
    \ENDFOR
    \STATE Sample a batch of $\{\mathbf{x}_i\}_{i=1}^B$ from the input matrix $(x_t)$
    \STATE $\theta\leftarrow Adam\bigg(\nabla_\theta\frac{1}{B}\sum_{i=1}^B\Big[-\mathbf{D}_\gamma(\mathbf{G}_\theta(\mathbf{x}_i))+$
    $\mu\lVert \mathbf{H}_{\eta}(\mathbf{G}_\theta(\mathbf{x}_i))- \mathbf{x}_i\rVert_2\Big],\alpha,\beta_1,\beta_2\bigg)$
    \STATE $\eta\leftarrow Adam\bigg(\nabla_\eta\frac{1}{B}\sum_{i=1}^B\left[\mu\lVert \mathbf{H}_{\eta}(\mathbf{G}_\theta(\mathbf{x}_i))- \mathbf{x}_i\rVert_2\right],$ $\alpha,\beta_1,\beta_2\bigg)$
   \ENDWHILE
\end{algorithmic}
\end{algorithm}

\section{Neural Network Parameter}
All the neural networks (encoder, decoder and discriminator) in the paper had three hidden layers, with the 100, 50, 25 neurons respectively. The input dimension for the generator was chosen such that $n=3m$. In the paper, $m=20$ was used for synthetic case, and $m=100$ for real data cases. The encoder and decoder both used hyperbolic tangent activation. The first two layers of the discriminator adopted hyperbolic tangent activation, and the last one linear activation.

The tuning parameter was chosen to be the same for all synthetic cases, with $\mu=0.1$, $\lambda=5$, $\alpha=0.0002$, $\beta_1=0.9$, $\beta_2=0.999$.  For the two real data cases, the hyper-parameters were set to be $\mu=0.01$, $\lambda=3$, $\alpha=0.001$, $\beta_1=0.9$, $\beta_2=0.999$.

\end{document}